\documentclass[letterpaper, 10 pt, conference]{ieeeconf}  

\IEEEoverridecommandlockouts                              

\usepackage{cite}
\usepackage{amsmath,amssymb,amsfonts}
\usepackage{algorithmic}
\usepackage{graphicx}
\usepackage{textcomp}
\usepackage{cuted}
\usepackage{capt-of}
\usepackage{float}
\usepackage{booktabs}
\usepackage{threeparttable}
\usepackage{mathtools}
\usepackage[mathscr]{eucal}
\usepackage[table,xcdraw]{xcolor}
\usepackage{colortbl}
\usepackage{colortbl}
\usepackage{balance}
\usepackage{color}

\usepackage{marvosym}
\usepackage{algorithmic}
\usepackage{indentfirst}
\usepackage{makecell}

\definecolor{car}{rgb}{0.39215686, 0.58823529, 0.96078431}
\definecolor{bicycle}{rgb}{0.39215686, 0.90196078, 0.96078431}
\definecolor{motorcycle}{rgb}{0.11764706, 0.23529412, 0.58823529}
\definecolor{truck}{rgb}{0.31372549, 0.11764706, 0.70588235}
\definecolor{other-vehicle}{rgb}{0.39215686, 0.31372549, 0.98039216}
\definecolor{person}{rgb}{1.        , 0.11764706, 0.11764706}
\definecolor{bicyclist}{rgb}{1.        , 0.15686275, 0.78431373}
\definecolor{motorcyclist}{rgb}{0.58823529, 0.11764706, 0.35294118}
\definecolor{road}{rgb}{1.        , 0.        , 1.        }
\definecolor{parking}{rgb}{1.        , 0.58823529, 1.        }
\definecolor{sidewalk}{rgb}{0.29411765, 0.        , 0.29411765}
\definecolor{other-ground}{rgb}{0.68627451, 0.        , 0.29411765}
\definecolor{building}{rgb}{1.        , 0.78431373, 0.        }
\definecolor{fence}{rgb}{1.        , 0.47058824, 0.19607843}
\definecolor{vegetation}{rgb}{0.        , 0.68627451, 0.        }
\definecolor{trunk}{rgb}{0.52941176, 0.23529412, 0.        }
\definecolor{terrain}{rgb}{0.58823529, 0.94117647, 0.31372549}
\definecolor{pole}{rgb}{1.        , 0.94117647, 0.58823529}
\definecolor{traffic-sign}{rgb}{1.        , 0.        , 0.    }

\makeatletter
\newcommand{\car@semkitfreq}{3.92}
\newcommand{\bicycle@semkitfreq}{0.03}
\newcommand{\motorcycle@semkitfreq}{0.03}
\newcommand{\truck@semkitfreq}{0.16}
\newcommand{\othervehicle@semkitfreq}{0.20}
\newcommand{\person@semkitfreq}{0.07}
\newcommand{\bicyclist@semkitfreq}{0.07}
\newcommand{\motorcyclist@semkitfreq}{0.05}
\newcommand{\road@semkitfreq}{15.30}  %
\newcommand{\parking@semkitfreq}{1.12}
\newcommand{\sidewalk@semkitfreq}{11.13}  %
\newcommand{\otherground@semkitfreq}{0.56}
\newcommand{\building@semkitfreq}{14.1}  %
\newcommand{\fence@semkitfreq}{3.90}
\newcommand{\vegetation@semkitfreq}{39.3}  %
\newcommand{\trunk@semkitfreq}{0.51}
\newcommand{\terrain@semkitfreq}{9.17} %
\newcommand{\pole@semkitfreq}{0.29}
\newcommand{\trafficsign@semkitfreq}{0.08}
\newcommand{\semkitfreq}[1]{{\csname #1@semkitfreq\endcsname}}

\makeatletter
\let\NAT@parse\undefined
\makeatother
\usepackage{hyperref}

\title{\LARGE \bf
MonoOcc: Digging into Monocular Semantic Occupancy Prediction
}

\author{Yupeng Zheng$^{1,2*}$, Xiang Li$^{3*}$, Pengfei Li$^{3}$, Yuhang Zheng$^{3}$, \\
 Bu Jin$^{1,2}$, Chengliang Zhong$^{3}$, Xiaoxiao Long$^{4\dagger}$, Hao Zhao$^{3}$, and Qichao Zhang$^{1,2\dagger}$\textsuperscript{\Letter}
\thanks{$^{1}$The State Key Laboratory of Multimodal Artificial Intelligence Systems, Institute of Automation, Chinese Academy of Sciences, Beijing 100190, China, \{zhangqichao2014, zhengyupeng2022\}@ia.ac.cn\ }
\thanks{$^{2}$School of Artificial Intelligence, University of Chinese Academy of Sciences, Beijing, China,}%
\thanks{$^{3}$Institute for AI Industry Research (AIR), Tsinghua University, China,}
\thanks{$^{4}$Department of Computer Science, the University of Hong Kong.}%
\thanks{$*$ Equal contribution.}
\thanks{$\dagger$ Project leader.}
\thanks{\textsuperscript{\Letter} Corresponding to zhangqichao2014@ia.ac.cn}
}

\begin{document}

\maketitle
\thispagestyle{empty}
\pagestyle{empty}

\begin{abstract}

Monocular Semantic Occupancy Prediction aims to infer the complete 3D geometry and semantic information of scenes from only 2D images. 
It has garnered significant attention, particularly due to its potential to enhance the 3D perception of autonomous vehicles. 
However, existing methods rely on a complex cascaded framework with relatively limited information to restore 3D scenes, including a dependency on supervision solely on the whole network's output, single-frame input, and the utilization of a small backbone. These challenges, in turn, hinder the optimization of the framework and yield inferior prediction results, particularly concerning smaller and long-tailed objects.
To address these issues, we propose MonoOcc. In particular, we (i) improve the monocular occupancy prediction framework by proposing an auxiliary semantic loss as supervision to the shallow layers of the framework and an image-conditioned cross-attention module to refine voxel features with visual clues, and (ii) employ a distillation module that transfers temporal information and richer knowledge from a larger image backbone to the monocular semantic occupancy prediction framework with low cost of hardware. 
With these advantages, our method yields state-of-the-art performance on the camera-based SemanticKITTI Scene Completion benchmark. Codes and models can be accessed at \href{https://github.com/ucaszyp/MonoOcc}{https://github.com/ucaszyp/MonoOcc}.

\end{abstract}
\section{Introduction}
3D scene understanding serves as a foundation for autonomous driving systems, exerting a direct influence on downstream tasks such as planning, navigation, VR\cite{yan2023distortion, yan2022multi}, map construction\cite{Wang2020Learningbased3O,Popovic2021VolumetricOM}. The past years have witnessed the rapid development and significant impact of lidar-based algorithms\cite{li2023lode, xia2023scpnet, chen2020boost, chen2018multi, li2021bifnet} in outdoor 3D scene understanding. Nevertheless, they are often considered expensive in terms of hardware for autonomous vehicles. Consequently, monocular scene understanding\cite{cao2022monoscene, Zheng2023STEPSJS, godard2019digging, Jin2023ADAPTAD, yan2023desnet, yan2022rignet} has garnered considerable attention from the robotics community due to its cost-efficiency and rich visual information. A popular topic in this domain is Semantic Occupancy Prediction, also denoted as Semantic Scene Completion (SSC) \cite{behley2019semantickitti}. Its objective is to predict the semantic occupancy of each voxel throughout the entire scene, encompassing both visible and occluded regions, while relying solely on monocular observations.


\begin{figure}
  \centering
  \includegraphics[width=0.48\textwidth]{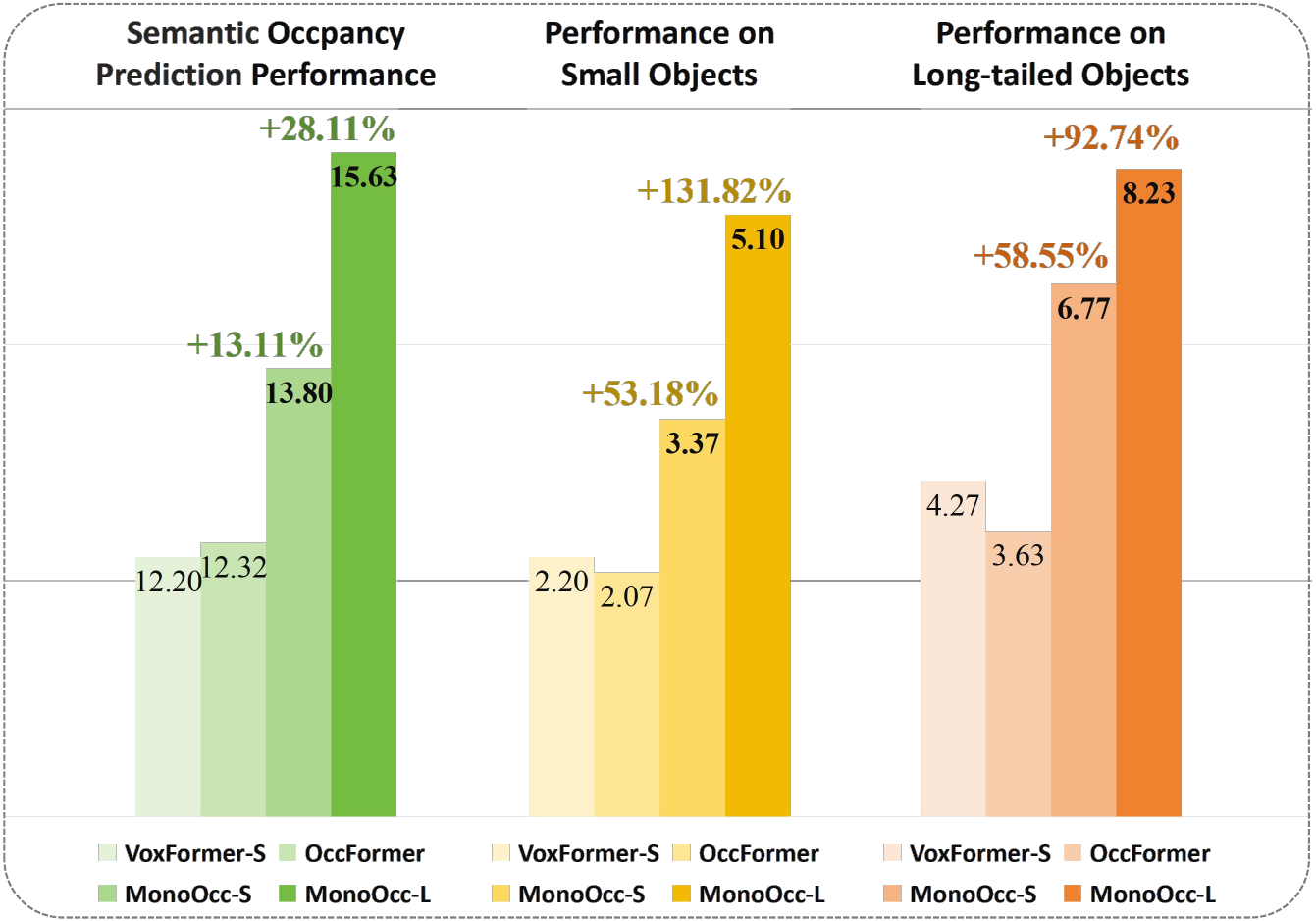}
  \caption{Quantitative results of semantic occupancy prediction on SemanticKITTI\cite{behley2019semantickitti} test set compared with the state-of-the-art VoxFormer-S\cite{li2023voxformer} and OccFormer\cite{zhang2023occformer}. Note that our method outperforms the latter methods in the SSC mIoU, while also achieving a significant boost on both small objects (\emph{bicycle}, \emph{motorcycle}, \emph{traffic-sign}) and long-tailed objects (\emph{truck}, \emph{other-vehicle}, \emph{other-ground}). Compared with VoxFormer-S, the relative percentage increase of our method on \textcolor[RGB]{88, 142, 49}{average performance}, \textcolor[RGB]{181, 139, 1}{small objects} and \textcolor[RGB]{198, 95, 16}{long-tailed objects} are denoted by \textcolor[RGB]{88, 142, 49}{\textbf{green}}, \textcolor[RGB]{181, 139, 1}{\textbf{yellow}} and \textcolor[RGB]{198, 95, 16}{\textbf{orange}}, respectively.}
  \vspace{-4mm}
  \label{fig:teaser}
\end{figure}
The existing cutting-edge method, VoxFormer\cite{li2023voxformer}, proposes a sparse-to-dense architecture, which aggregates 2D features to voxel space with depth-based queries and completes the entire scene conditioned on the feature of queries.
It should be emphasized that the inaccuracies of depth estimation affect the accuracy of the query, leading to an increase in the difficulty of subsequent completion and semantic parsing. Besides, all of the existing methods such as \cite{cao2022monoscene,li2023voxformer, wei2023surroundocc} rely solely on the supervision of 3D ground truth to train the deep cascaded architecture, including the 2D image backbone, 2D-3D view transformer, and 3D completion network, bringing challenges to the optimization of heterogeneous sub-modules.
Additionally, these methods only utilize visual information from a single frame input, resulting in poor performance, particularly causing failures in predictions for small objects and long-tailed objects (see Fig.\ref{fig:teaser}).

To this end, we first propose an image-conditioned cross-attention module, aiming to refine inaccurate voxel features brought by depth estimation with the extra information from image features and introduce an auxiliary semantic loss as supervision to the shallow layers of the small framework, facilitating more efficient optimization.
Secondly, we employ a large pre-trained image backbone instead of small models trained on benchmark datasets as the former arts used to assist in SSC. 
Recent research (e.g., \cite{wang2023internimage, liu2022swin}) has demonstrated that large image backbones can significantly enhance the adaptability and generality of 2D image semantic segmentation.
However, how to efficiently utilize these models in the SSC task has not been explored. Considering the efficiency and resource constraints in real-world applications, we propose to use model distillation to get more compact yet efficient models that can approximate the behavior of larger models. 
As shown in Fig. \ref{fig:main}, we denote the larger model as a privileged branch, inspired by the idea of privileged learning which is widely recognized in the robotics community\cite{chen2020learning, lee2020learning}. 
This branch is designed to take temporal image frames as inputs, thus mitigating uncertainty in occluded areas. 
As illustrated in Fig. \ref{fig:teaser}, the comparison between SOTAs' SSC results and ours on SemanticKITTI \cite{behley2019semantickitti} test set demonstrates that our method achieves a significant gain on general, small and long-tailed objects.

For easy reference, we summarize our contributions below.
\begin{itemize}
\item[$\bullet$] We propose an image-conditioned cross-attention module and semantic auxiliary loss to improve the performance of Monocular SSC.
\item[$\bullet$] We propose a privileged branch with pre-training a larger backbone and employing a cross-view transformer to acquire more visual cues from temporal frames.
\item[$\bullet$] We propose a distillation module to transfer knowledge from the privileged branch to the monocular branch.
\item[$\bullet$] We achieve SOTA performance on SemanticKITTI benchmark\cite{behley2019semantickitti} and release our codes and models.
\end{itemize}

\section{Related Works}

\textbf{Camera-based 3D Perception.} In recent years, there has been a growing interest in camera-based 3D sensing techniques\cite{li2022bevformer, huang2021bevdet, li2023bevdepth, saha2022translating}, primarily driven by the advantages of richer visual information, ease of deployment, and cost-effectiveness offered by cameras. Recent research in camera-based 3D perception focuses on constructing BEV feature representations and subsequently performing various downstream tasks in the BEV space. The Lift-Splat-Shoot (LSS) method\cite{philion2020lift}, along with its subsequent advancements\cite{hu2021fiery, li2023bevdepth}, serves as the archetypal technique for forward projection. LSS projects image features into 3D space and aggregates them into the BEV space, incorporating depth uncertainty through predicted pixel-wise depth distributions. BEVFormer\cite{li2022bevformer} represents one of the backward projection methods, utilizing deformable attention-based spatiotemporal transformers to construct BEV queries and aggregate corresponding 2D features from multiple frames into the BEV space. Given that 3D occupancy representation contains richer spatial information compared to BEV representation, it plays a crucial role in the perception and planning abilities of self-driving cars. Consequently, there is a noticeable shift towards employing camera-based solutions in 3D Semantic Occupancy Prediction. 

\textbf{Camera-based 3D Semantic Occupancy Prediction}.
 After the introduction of SemanticKITTI dataset\cite{behley2019semantickitti}, an abundance of outdoor Single-View 3D Semantic Occupancy Prediction (synonymous with Semantic Scene Completion) techniques have emerged. MonoScene\cite{cao2022monoscene} is the pioneering method for monocular semantic occupancy prediction, which proposes 2D-3D feature projections along the line of sight to generate voxel features and utilizes the 3D UNet to process the volumetric data. TPVFormer\cite{huang2023tri} introduces a simple yet efficient tri-perspective view representation as an alternative to the BEV representation, enabling the capture of 3D structural intricacies. OccFormer\cite{zhang2023occformer} devises dual-path transformer blocks comprising local and global transformers to decompose the 3D processing. VoxFormer\cite{li2023voxformer} replaces BEV queries with depth-based proposal queries to aggregate features from images and introduces an MAE-like design to achieve dense occupancy completion with sparse queries.

\section{Method}\label{sec:method}
\begin{figure*}
  \centering
  \includegraphics[width=0.95\textwidth]{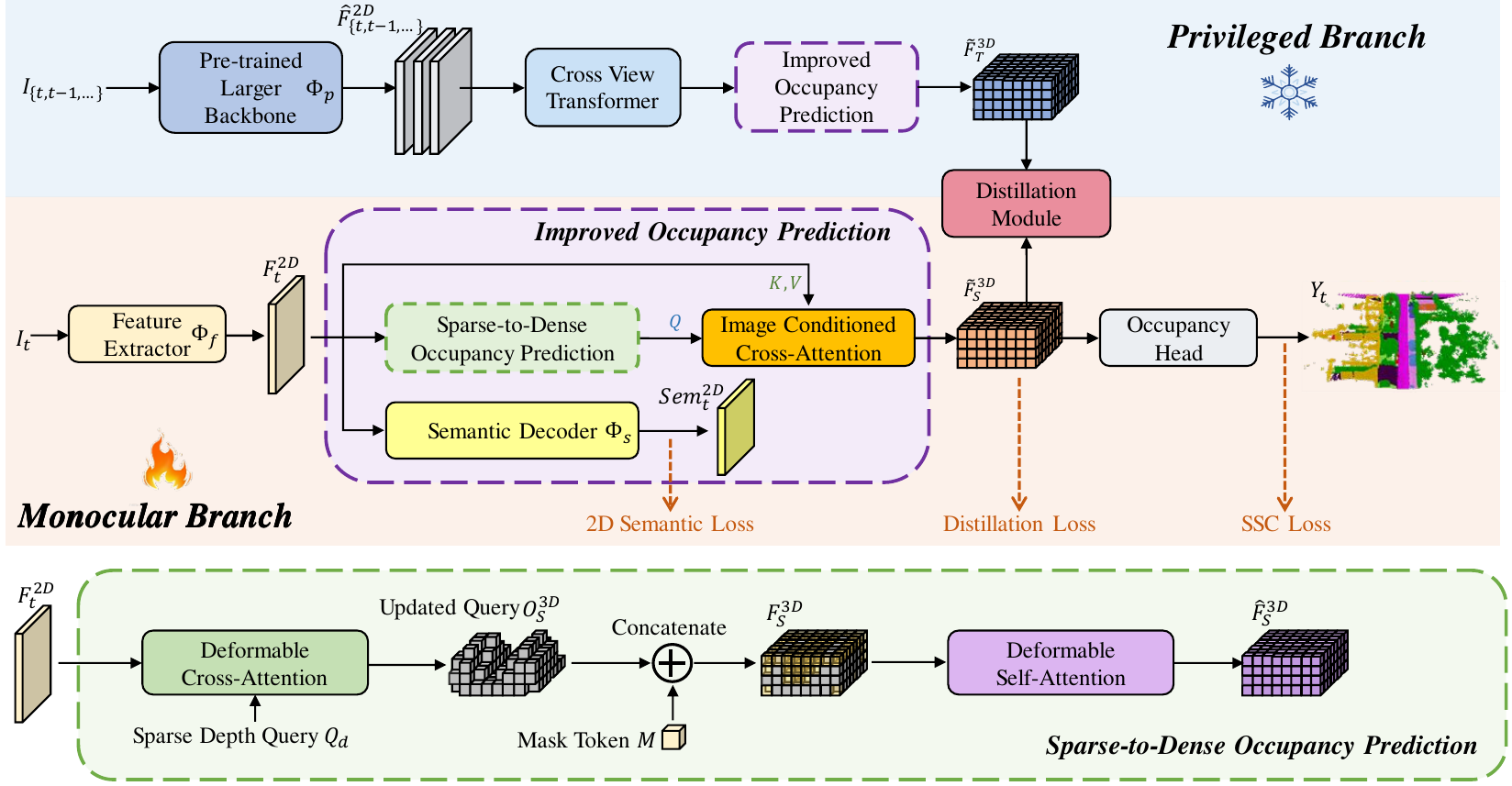}
  \vspace{-0.3cm} 
  \caption{The architecture of our proposed framework (see section \ref{sec:method} for details)  
  }
  \vspace{-0.5cm} 
  \label{fig:main}
\end{figure*}
An overall framework of MonoOcc is illustrated in Fig. \ref{fig:main}

We briefly describe the sparse-to-dense monocular 3D semantic occupancy prediction pipeline of a  baseline method in section \ref{A}.
Two innovations, including an image-conditioned cross attention and a 2D semantic auxiliary loss, are proposed to improve the current framework in section \ref{B}.
To promote the performance of small objects and long-tailed objects, we further propose a privileged branch by pre-training a larger image backbone and introducing a cross view transformer to enhance temporal view features in section \ref{C}.
Finally, we propose a distillation module to transfer the knowledge from the privileged branch to the monocular branch, making a trade-off between performance and efficiency in section \ref{D}.

\subsection{\textbf{Sparse-to-dense Monocular 3D Semantic Occupancy Prediction}}\label{A}
\textbf{Image Feature Extractor.} To extract 2D feature maps $F_t^{2D} \in \mathbb{R}^{d\times h\times w}$ from corresponding RGB images $I_{t}$, an image feature extractor $\Phi_f$ is constructed by employing ResNet-50\cite{he2016deep} as backbone and FPN\cite{lin2017feature} as neck, where $d$ and $(h, w)$ represent the dimension and resolution of the image feature, respectively. 
Later we will leverage a stronger image feature extractor pre-trained on a bunch of diverse autonomous driving datasets.

\textbf{Depth-based Query.} Following VoxFormer\cite{li2023voxformer}, we generate a total of $N_d$ queries $Q_d$ based on the depth map predicted by a pre-trained depth network. Specifically, we utilize pixel-wise depth to unproject pixels into 3D space, and then obtain initial occupancy by voxelizing these points. Afterward, we acquire a tractable number of basically reasonable initial queries $Q_{d}$ by correcting initial occupancy with an occupancy prediction network (LMSCNet\cite{roldao2020lmscnet}).

\textbf{Voxel Feature Generator.} Following VoxFormer\cite{li2023voxformer}, the process of generating voxel features $\hat{F}_{S}^{3D} \in \mathbb{R}^{x\times y\times z\times d}$ with resolution $(x, y, z)$  can be divided into two steps:

1) We acquire $O_{S}^{3D} \in \mathbb{R}^{N_d \times d}$, the feature of visible regions, by utilizing $Q_d$ to aggregate 2D feature $F_t^{2D}$ into 3D space with deformable cross-attention mechanism (DCA)\cite{zhu2020deformable}: 
\begin{equation}
    O_{S}^{3D} = {\rm {DCA}} \left(Q_d, F_t^{2D}\right).
\end{equation}
For a set of temporal view features ${F}_{\{t, t-1, ...\}}^{2D} \in \mathbb{R}^{N \times d\times h\times w}$ with a valid quantity of $N$, $O_{T}^{3D}$ is obtained by an average of aggregated feature:
\begin{equation}
    O_{T}^{3D} = \frac{1}{N}{\rm {DCA}} \left(Q_d, F_{\{t,t-1,...\}}^{2D}\right).
\end{equation}

2) We acquire initial voxel features $F_S^{3D} \in \mathbb{R}^{x\times y\times z\times d}$ of the whole scene  by filling the occluded regions with mask token $M \in \mathbb{R}^{d} $ and then update $F_S^{3D}$ to $\hat{F}_S^{3D}$ with deformable self-attention mechanism (DSA)\cite{zhu2020deformable}:
\begin{equation}\label{eq:DSA}
    {\hat{F}_{S}^{3D}} = {\rm {DSA}} \left(F_S^{3D},F_S^{3D}\right).
\end{equation}

\textbf{Semantic Voxel Map}. The predicted semantic voxel map $Y_t\in \mathbb{R}^{X\times Y\times Z\times C}$ is obtained by up-sampling and linear projection of $\hat{F}_S^{3D}$, where $(X, Y, Z)$ denotes the resolution of the 3D volume, $C$ represents the number of classes, including non-occupied.

\subsection{\textbf{Improved Architecture for Monocular Semantic Occupancy Prediction}}\label{B}

\textbf{Image-Conditioned Cross-Attention.} 
VoxFormer\cite{li2023voxformer} treats semantic occupancy prediction as a generative task. The MAE-like transformer hallucinates the occluded scene conditioned on the visible scene. Mathematically, given the features of the visible region $O_S^{3D}$ and the initial voxel feature $F_S^{3D}$, the refined features of the entire scene $\hat{F}_S^{3D}$ can be acquired by:
\begin{equation}
    \hat{F}_S^{3D} = {\rm Complete}\left(F_S^{3D} | O_S^{3D}\right),
\end{equation}
where $\rm Complete(\cdot | \cdot)$ means complete the former conditioned on the latter.
Reviewing the generation of $O_S^{3D}$, the inaccuracy of depth estimation introduces inaccurate geometric information, bringing uncertainty to the completion of the entire scene. We believe that the features from the input image can help correct the inaccuracies as they can provide semantic clues. Thus, we complete the scene conditioned on $O_S^{3D}$ and $F_t^{2D}$:
\begin{equation}
    \tilde{F}_S^{3D} = {\rm Complete}\left(F_S^{3D} | O_S^{3D}, F_t^{2D}\right){.}
\end{equation}
This can be naturally achieved through the deformable cross-attention mechanism. Specifically, we treat the $\hat{F}_S^{3D}$ as the query, $F_t^{2D}$ as the key and value, and leverage deformable cross attention to obtain the corrected image-conditioned voxel features from the refined feature:
\begin{equation}\label{eq:DCA}
    \tilde{F}_S^{3D} = {\rm {DCA}} \left(\hat{F}_S^{3D}, F_t^{2D}\right).
\end{equation}

\textbf{2D Semantic Auxiliary Loss.}
The occupancy prediction network is a long cascaded framework with components of different domains including a 2D feature extractor, 2D-3D cross-attention, 3D completion self-attention, and occupancy head, which increases difficulties of optimization. 
To address this problem, we propose a 2D auxiliary semantic loss as deep supervision to the feature extractor. It provides a shorter path for backpropagation, enabling better optimization of the feature extractor, which serves as the source of features for the entire framework. 

Regarding the implementation of 2D semantic loss, we first employ a semantic decoder $\Phi_s$ composed of convolutional layers and a fully connected layer to predict semantic map $Sem_t^{2D}$ from image feature $F_t^{2D}$:
\begin{equation}
    Sem_t^{2D} = \Phi_s\left(F_t^{2D}\right).
\end{equation}
Then we project point clouds with semantic labels to corresponding images to generate sparse ground truth. Finally, we employ cross-entropy loss $\mathcal{L}_{sem}$ between ground truth and $Sem_t^{2D}$ to directly optimize the feature extractor.

\subsection{\textbf{Privileged Branch}}\label{C}

\textbf{Pre-training Scaled-Up Feature Extractor.} Increasing the size of the model is an effective strategy to improve the accuracy of dense prediction tasks. However, due to the limited training samples in SemanticKITTI\cite{behley2019semantickitti} which only contains about 12K images, using a larger image backbone would lead to overfitting. 
Moreover, SemanticKITTI not only lacks dense semantic labels but also contains multiple long-tailed classes such as \emph{other-vehicle} (0.20\%), \emph{truck} (0.16\%), and \emph{other-ground} (0.56\%) which only have very limited label points as supervision, making it ineffective to train a larger backbone.

To cope with these two problems, we pre-train the larger image backbone with more data of the driving scenario. And we employ the InternImage-XL\cite{wang2023internimage} loading the pre-trained model as the visual backbone $\Phi_{p}$ for the privilege branch to extract 2D features from multiple frames of images:
\begin{equation}
    \hat{F}_{\{t, t-1, ...\}}^{2D} = \Phi_{p}\left(I_{\{t, t-1, ...\}}^{2D}\right) {,} 
\end{equation}
where ${t}$ represents the $t$-th frame image arranged in chronological order. More details about pre-training are elaborated in section \ref{details}.

\textbf{Cross View Transformer}. Previous work\cite{li2023voxformer,chen2018temporal, zhao2020spatial} has demonstrated that temporal information boosts downstream 3D scene perception. 
When aggregating 2D features into 3D with deformable cross-attention, only the centroid of voxels are projected to the image as reference points. 
Limited by the voxel resolution, the deviation between the real position of objects in 3D space and the voxel centroid can affect the occupancy prediction of a single frame, which can be alleviated by involving more viewpoints.
To acquire as much visual information as possible, we adopt multiple frames as the inputs of the privileged branch. 
 
To further enhance the features of temporal views, we introduce the Cross View Transformer (CVT)\cite{wei2023surrounddepth} to integrate knowledge across multi-views, which is proved to be effective on dense prediction tasks such as depth estimation\cite{wei2023surrounddepth}, optical flow estimation\cite{xu2022gmflow,xu2023unifying}, and map-view semantic segmentation\cite{zhou2022cross}. 
In particular, we first add positional encoding to the independently extracted temporal features. Then, we input pairs of adjacent features in chronological order into the CVT for feature enhancement. The enhanced features are lifted to the voxel space through the deformable cross-attention mechanism (DCA).

\subsection{\textbf{Distillation Module}}\label{D}
In the previous subsection, we adopt a temporal view as input and scale up the image backbone to acquire visual clues as rich as possible. 
However, the usage of multiple frames as input and the scaling up of the 2D backbone significantly increase computational costs, affecting deployment in autonomous driving systems. 
To address this, inspired by privilege learning\cite{lee2020learning}, we propose a distillation module, composed of a privileged teacher branch and a monocular student branch. 
The module aims to transfer the knowledge from the privileged branch, which has richer clues of temporal information and a larger backbone prior, to the monocular branch, resulting in performance improvement for the monocular branch. 
Specifically, inspired by Frustums Proportion Loss from MonoScene\cite{cao2022monoscene}, we use Kullback-Leibler Divergence as the loss function to provide the cues from teacher to student:

\begin{equation}
    \mathcal{L}_{distill} = {\rm KL}(\tilde{F}_T^{3D} || \tilde{F}_S^{3D}).
\end{equation}

\subsection{\textbf{Training Loss}}
Reviewing the training process, we employ multiple loss functions to supervise varying depths of the network. 
\begin{itemize}
    \item [$\bullet$] For the feature extractor, we introduce loss $\mathcal{L}_{sem}$.
    \item[$\bullet$] For the completion network, we propose temporal distillation loss $\mathcal{L}_{distill}$.
    \item [$\bullet$] For the final output semantic grid map, we utilize Loss functions $\mathcal{L}_{ssc}$, $\mathcal{L}_{scal}^{sem} $, and $\mathcal{L}_{scal}^{geo} $ from MonoScene\cite{cao2022monoscene}. 
\end{itemize}

The total loss function can be represented as follows:
\begin{equation}
\begin{split}
    \mathcal{L} &= \lambda_{1} {\mathcal{L}}_{sem} + \lambda_{2} {\mathcal{L}}_{distill} + \lambda_{3} \mathcal{L}_{ssc} \\ &+ \lambda_{4} \mathcal{L}_{scal}^{sem} + \lambda_{5} \mathcal{L}_{scal}^{geo},
\end{split}
\end{equation}
where $\lambda_{1}, \lambda_{2}, \lambda_{3}, \lambda_{4} {,}$ and $\lambda_{5}$ are hyper-parameters.

\section{Experiments}

\subsection{\textbf{Dataset}}

We evaluate our method on the SemanticKITTI dataset\cite{behley2019semantickitti}, which provides dense semantic occupancy annotations of all lidar scans from the KITTI Odometry Benchmark\cite{geiger2012we}. Each lidar scan of SemanticKITTI covers a range of $[0\sim 51.2\rm m, -25.6\sim 25.6\rm m, -2\sim 4.4\rm m]$ ahead of the ego car. The ground-truth semantic occupancy is represented as $256\times 256\times 32$ 3D voxel grids through voxelizing aggregated lidar scans with 0.2m resolution. Each voxel is annotated with 20 classes (19 semantic classes and 1 free). 
The official split for train, validation, and test sets is employed. We report our main result (Table \ref{table:kitti_test}) on the test set and do ablation studies (Table \ref{tab:ablation1}, \ref{tab:ablation2}) on the validation set.

\textbf{Evaluation metrics.} Following common practices, we report the mean intersection over union (mIoU) of 19 semantic classes for the Semantic Occupancy Prediction task.

\begin{table*}
    \scriptsize
    \setlength{\tabcolsep}{0.005\linewidth}
    \setlength{\tabcolsep}{0.004\linewidth}
    \caption{\textbf{Semantic scene completion results on SemanticKITTI~\cite{behley2019semantickitti} test set.}}
    \vspace{-2mm}
    \newcommand{\classfreq}[1]{{~\tiny(\semkitfreq{#1}\%)}}  %
    \centering
    \begin{tabular}{l|c|c|c c c c c c c c c c c c c c c c c c c|c}
        \toprule
        & & & \multicolumn{20}{c}{Semantic Occupancy Prediction} \\
        Method & Pub & Input & \rotatebox{90}{\textcolor{road}{$\blacksquare$} road\classfreq{road}} 
        & \rotatebox{90}{\textcolor{sidewalk}{$\blacksquare$} sidewalk\classfreq{sidewalk}}
        & \rotatebox{90}{\textcolor{parking}{$\blacksquare$} parking\classfreq{parking}} 
        & \rotatebox{90}{\textcolor{other-ground}{$\blacksquare$} other-ground\classfreq{otherground}} 
        & \rotatebox{90}{\textcolor{building}{$\blacksquare$} building\classfreq{building}} 
        & \rotatebox{90}{\textcolor{car}{$\blacksquare$} car\classfreq{car}} 
        & \rotatebox{90}{\textcolor{truck}{$\blacksquare$} truck\classfreq{truck}} 
        & \rotatebox{90}{\textcolor{bicycle}{$\blacksquare$} bicycle\classfreq{bicycle}} 
        & \rotatebox{90}{\textcolor{motorcycle}{$\blacksquare$} motorcycle\classfreq{motorcycle}} 
        & \rotatebox{90}{\textcolor{other-vehicle}{$\blacksquare$} other-vehicle\classfreq{othervehicle}} 
        & \rotatebox{90}{\textcolor{vegetation}{$\blacksquare$} vegetation\classfreq{vegetation}} 
        & \rotatebox{90}{\textcolor{trunk}{$\blacksquare$} trunk\classfreq{trunk}} 
        & \rotatebox{90}{\textcolor{terrain}{$\blacksquare$} terrain\classfreq{terrain}} 
        & \rotatebox{90}{\textcolor{person}{$\blacksquare$} person\classfreq{person}} 
        & \rotatebox{90}{\textcolor{bicyclist}{$\blacksquare$} bicyclist\classfreq{bicyclist}} 
        & \rotatebox{90}{\textcolor{motorcyclist}{$\blacksquare$} motorcyclist\classfreq{motorcyclist}} 
        & \rotatebox{90}{\textcolor{fence}{$\blacksquare$} fence\classfreq{fence}} 
        & \rotatebox{90}{\textcolor{pole}{$\blacksquare$} pole\classfreq{pole}} 
        & \rotatebox{90}{\textcolor{traffic-sign}{$\blacksquare$} traffic-sign\classfreq{trafficsign}} 
        & mIoU\\
        \midrule
        LMSCNet*~\cite{roldao2020lmscnet} & 3DV 2020 & Camera & 46.70 & 19.50 & 13.50 & 3.10 & 10.30 & 14.30 & 0.30 & 0.00 & 0.00 & 0.00 & 10.80 & 0.00 & 10.40 & 0.00 & 0.00 & 0.00 & 5.40 & 0.00 & 0.00 & 7.07\\ %
        3DSketch*~\cite{chen20203d} & CVPR 2020 & Camera & 37.70 & 19.80 & 0.00 & 0.00 & 12.10 & 17.10 & 0.00 & 0.00 & 0.00 & 0.00 & 12.10 & 0.00 & 16.10 & 0.00 & 0.00 & 0.00 & 3.40 & 0.00 & 0.00 & 6.23 \\ %
        AICNet*~\cite{li2020anisotropic} & CVPR 2020 & Camera & 39.30	& 18.30 & 19.80 & 1.60 & 9.60	& 15.30	& 0.70	& 0.00	& 0.00	& 0.00	& 9.60	& 1.90	& 13.50	& 0.00	& 0.00	& 0.00	& 5.00	& 0.10	& 0.00 & 7.09 \\ %
        JS3C-Net*~\cite{yan2021sparse} & AAAI 2021 & Camera & 47.30 & 21.70 & 19.90 & 2.80 & 12.70 & 20.10 & 0.80 & 0.00 & 0.00 & 4.10 & 14.20 & 3.10 & 12.40 & 0.00 & 0.20 & 0.20 & 8.70 & 1.90 & 0.30 & 8.97 \\
        MonoScene~\cite{cao2022monoscene} & CVPR 2022 & Camera & 54.70 & 27.10 & 24.80 & 5.70 & 14.40 & 18.80 & 3.30 & 0.50 & 0.70 & 4.40 & 14.90 & 2.40 & 19.50 & 1.00 & 1.40 & \underline{0.40} & 11.10 & 3.30 & 2.10 & 11.08 \\
        TPVFormer~\cite{huang2023tri} & CVPR 2023 & Camera & 55.10 & 27.20 & 27.40 & 6.50 & 14.80 & 19.20 & 3.70 & 1.00 & 0.50 & 2.30 & 13.90 & 2.60 & 20.40 & 1.10 & 2.40 & 0.30 & 11.00 & 2.90 & 1.50 & 11.26 \\
        VoxFormer-S~\cite{li2023voxformer} & CVPR 2023 & Camera & 53.90 & 25.30 & 21.10 & 5.60 & 19.80 & 20.80 & 3.50 & 1.00 & 0.70 & 3.70 & 22.40 & 7.50 & 21.30 & 1.40 & \underline{2.60} & 0.00 & 11.10 & 5.10 & 4.90 & 12.20 \\
        VoxFormer-T$\dagger$~\cite{li2023voxformer} & CVPR 2023 & Camera & 54.10 & 26.90 & 25.10 & 7.30 & \textbf{23.50} & 21.70 & 3.60 & 1.90 & 1.60 & 4.10 & \underline{24.40} & 8.10 & \underline{24.20} & 1.60 & 1.10 & 0.00 & 13.10 & \underline{6.60} & 5.70 & 13.41 \\
        OccFormer~\cite{zhang2023occformer} & ICCV 2023 & Camera & 55.90 & \underline{30.30} & \textbf{31.50} & 6.50 & 15.70 & 21.60 & 1.20 & 1.50 & \underline{1.70} & 3.20 & 16.80 & 3.90 & 21.30 & \underline{2.20} & 1.10 & 0.20 & 11.90 & 3.80 & 3.70 & 12.32 \\
        SurroundOcc~\cite{wei2023surroundocc} & ICCV 2023 &  Camera & \underline{56.90} & 28.30 & \underline{30.20} & 6.80 & 15.20 & 20.60 & 1.40 & 1.60 & 1.20 & 4.40 & 14.90 & 3.40 & 19.30 & 1.40 & 2.00 & 0.10 & 11.30 & 3.90 & 2.40 & 11.86 \\
        \midrule
        MonoOcc-S(Ours) & ICRA 2024 & Camera & 55.20 & 27.80 & 25.10 & \underline{9.70} & 21.40 & \underline{23.20} & \underline{5.20} & \underline{2.20} & 1.50 & \underline{5.40} & 24.00 & \underline{8.70} & 23.00 & 1.70 & 2.00 & 0.20 & \underline{13.40} & 5.80 & \underline{6.40} & \underline{13.80} \\
        MonoOcc-L(Ours) & ICRA 2024 & Camera & \textbf{59.10} & \textbf{30.90} & 27.10 & \textbf{9.80} & \underline{22.90} & \textbf{23.90} & \textbf{7.20} & \textbf{4.50} & \textbf{2.40} & \textbf{7.70} & \textbf{25.00} & \textbf{9.80} & \textbf{26.10} & \textbf{2.80} & \textbf{4.70} & \textbf{0.60} & \textbf{16.90} & \textbf{7.30} & \textbf{8.40} & \textbf{15.63} \\
        \bottomrule
        \multicolumn{20}{l}{* represents the results adapted for RGB inputs and reported in MonoScene~\cite{cao2022monoscene}.}\\
        \multicolumn{20}{l}{$\dagger$ represents the result with temporal inputs.}\\
        \multicolumn{20}{l}{The best and second-best performances are represented by \textbf{bold} and \underline{underline} respectively.}
    \end{tabular}\\
    \label{table:kitti_test}
    \vspace{-4mm}
\end{table*}

\begin{table}
    \centering
    \setlength{\tabcolsep}{0.015\linewidth}
    \caption{Ablation study on the effectiveness of each proposed component on improved monocular branch (Validation SET).}
    \vspace{-2mm}
    \begin{tabular}[b]{cccc|cc}
        \toprule
        row & 
        \makecell{2D Semantic \\ Auxiliary Loss} & 
        \makecell{Distillation \\ Module} & 
        \makecell{Image Conditioned \\ Cross Attn.} & 
        \makecell{Train \\ MEM} &
        mIoU$ \uparrow$\\
        \midrule
        1 & $\times$ & $\times$ & $\times$ & 16G & 12.35 \\
        2 & \checkmark & $\times$ & $\times$ & 16G & 13.08 \\
        3 & $\times$ & \checkmark (L) & $\times$ & 16G & 12.88 \\
        4 & $\times$ & $\times$ & \checkmark & 18G & 12.70 \\
        5 & \checkmark & \checkmark (L) & $\times$ & 16G & 13.26 \\
        6 & \checkmark & $\times$ & \checkmark & 18G & 13.14 \\
        7 & $\times$ & \checkmark(L) & \checkmark & 18G & 13.35 \\
        8 & \checkmark & \checkmark (S) & \checkmark & 18G & 13.80 \\
        9 & \checkmark & \checkmark (L) & \checkmark & 18G &\textbf{14.01} \\
        \bottomrule
    \end{tabular}
    \vspace{-6mm}
    \label{tab:ablation1}
\end{table}

\subsection{\textbf{Implementation Details}}\label{details}
We provide two versions of MonoOcc, namely MonoOcc-S and MonoOcc-L. As shown in Fig. \ref{fig:main}, MonoOcc-S employs ResNet50 as the image backbone of the monocular branch, while MonoOcc-L replaces the ResNet50 with our pre-trained larger backbone. For training, we set the hyper-parameters as follows: $\lambda_{1} = 4.0$, $\lambda_{2} = 3.0$, $\lambda_{3} = 2.0$, $\lambda_{4} = 1.0$ and $\lambda_{5} = 0.5$. We train MonoOcc-S on 4 GeForce 3090 GPUs and MonoOcc-L on 4 A100 GPUs for 20 epochs.

To pre-train the larger backbone, we choose the InternImage-XL\cite{wang2023internimage}(350M parameters) as our backbone, and we process approximately 200K training data from open-source autonomous driving datasets including Mapillary Vistas\cite{neuhold2017mapillary}, KITTI-360\cite{liao2022kitti}, BDD100K\cite{yu2020bdd100k}, Cityscapes\cite{cordts2016cityscapes}, and nuImages\cite{caesar2020nuscenes}. Based on the open-source pre-trained model of InternImage-XL, we first train on the Mapillary Vistas dataset, which includes 124 semantic categories with detailed annotations of road elements and long-tailed objects, effectively enhancing the model's understanding of the driving scenario. 
Then, we semantically align the KITTI-360, BDD100K, Cityscapes, and nuImages datasets, further training on 19 common road elements specific to driving scenarios. 
Finally, We pre-train the backbone on 8 A100 GPUs for a total of 20k iterations.  

\subsection{\textbf{Quantitative and Qualitative Results}}
\begin{figure*}
  \centering
  \includegraphics[width=0.95\textwidth]{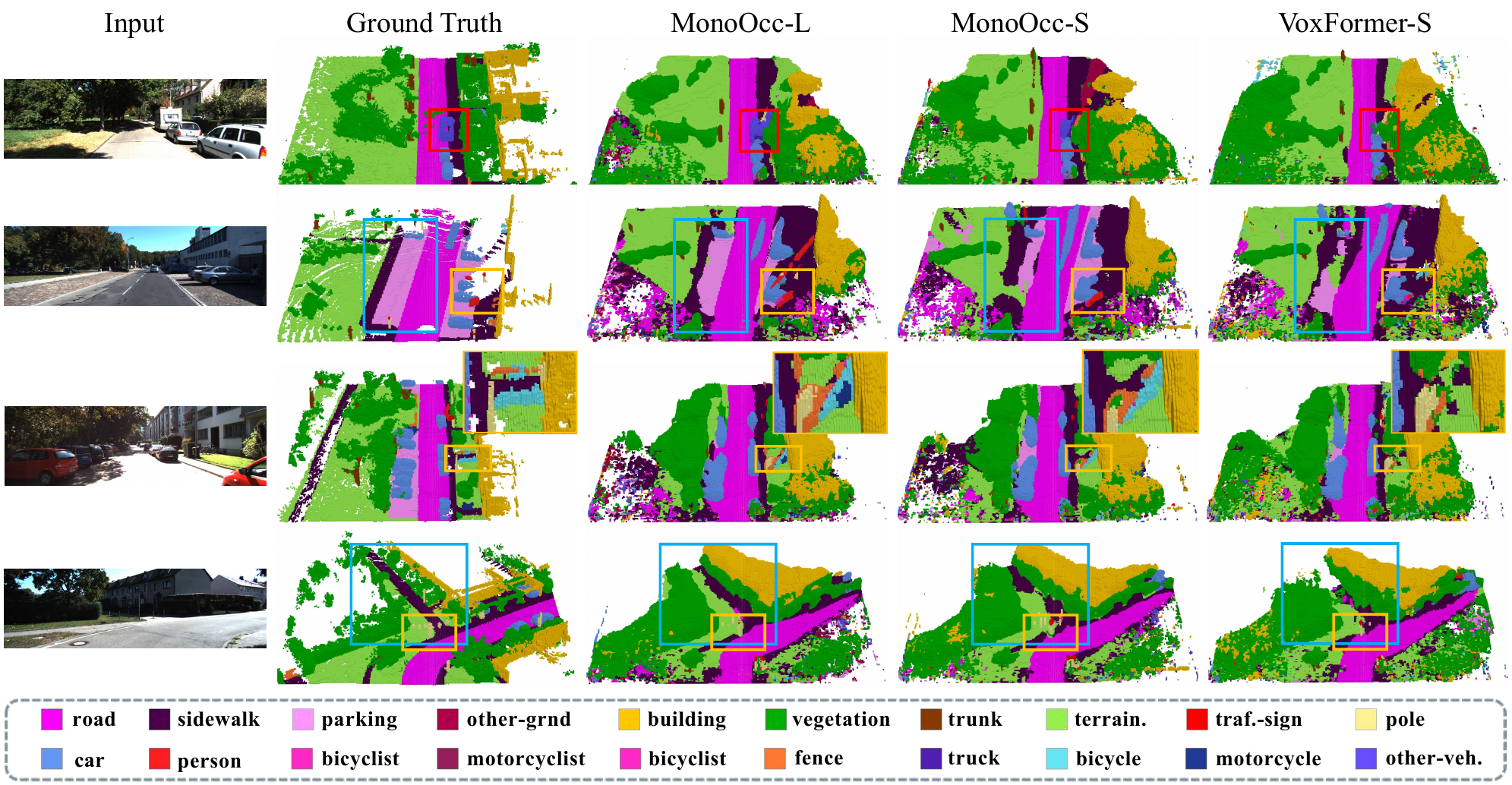}
  \vspace{-0.2cm} 
  \caption{Qualitative results of our method and VoxFormer on SemanticKITTI dataset}
  \vspace{-0.3cm} 
  \label{fig:vis}
\end{figure*}

In this section, we compare our method with competitive baselines on the test split of SemanticKITTI in Table \ref{table:kitti_test} and demonstrate qualitative results in Fig. \ref{fig:vis}.
Table \ref{table:kitti_test} shows the comparison results with some methods adapted for RGB input such as LMSCNet\cite{roldao2020lmscnet}, JS3C-Net\cite{yan2021sparse} and other competitive camera-based semantic occupancy prediction methods such as TPVFormer\cite{huang2023tri} VoxFormer\cite{li2023voxformer}, and OccFormer\cite{zhang2023occformer} on SemanticKITTI.
Overall, our method achieves a significant improvement over the other baselines in nearly all classes and sets new SOTA. To be specific, MonoOcc-S and MonoOcc-L achieve a remarkable boost of 1.60 mIoU and 3.43 mIoU, respectively, compared to the baseline method VoxFormer\cite{li2023voxformer}. For the sake of fairness, in the following, we compare MonoOcc-S with VoxFormer-S to analyze the advantages of our methods.
Delving into the qualitative results, we find that our algorithm achieves the expected performance improvement on long-tailed objects and small objects.

\textbf{Our Method performs better on long-tailed objects.} As shown in Table \ref{table:kitti_test}, MonoOcc-S shows a significant improvement in predicting long-tailed objects compared with VoxFormer-S, such as the \emph{other-ground} (0.56\%, 5.60 $\to$ 9.70), \emph{other-vehicle} (0.20\%, 3.70 $\to$ 5.40) and \emph{truck} (0.16\%, 3.50 $\to$ 5.20).

\textbf{Our Method performs better on small objects.} As shown in Table \ref{table:kitti_test}, MonoOcc-S demonstrates a significant boost in predicting small objects compared with VoxFormer-S, such as the \emph{bicycle} (1.00 $\to$ 2.20), \emph{motorcycle} (0.70 $\to$ 1.50) and \emph{traffic-sign} (4.90 $\to$ 6.40).

\textbf{Qualitative Results.} To demonstrate the performance of our algorithm more intuitively, we also provide the qualitative visualization of the predicted semantic occupancy in Fig. \ref{fig:vis}. 
The four rows demonstrate the superiority of our method on \textcolor[RGB]{255,0,0}{\textbf{long-tailed objects}}, \textcolor[RGB]{255,192,0}{\textbf{small objects}} and \textcolor[RGB]{0,176,240}{\textbf{road segmentation}}.
The \textcolor[RGB]{255,0,0}{red box} in the first row shows that VoxFormer-S cannot estimate the instance of \emph{other-vehicle}. 
The \textcolor[RGB]{255,192,0}{orange box} in the second row, third row, and last row show that our method performs better on small objects like \emph{person}, \emph{bicycle} and \emph{pole}, respectively. 
To further demonstrate the effectiveness of our method, we show the impressive prediction result of \emph{road}, a crucial category for autonomous vehicles to estimate the drivable area, within \textcolor[RGB]{0,176,240}{blue box} in the second row and last row.

\begin{table}
    \centering
    \setlength{\tabcolsep}{0.03\linewidth}
    \caption{Ablation study on the effectiveness of scaling up and pre-training  (Validation SET).}
    \vspace{-2mm}
    \begin{tabular}[b]{cc|cc}
        \toprule
        \makecell{Scaling-up} & 
        \makecell{Pre-training} & 
        \makecell{Test MEM} & mIoU$ \uparrow$\\
        \midrule
        $\times$ & $\times$ & 8G & 12.35 \\
        \checkmark & $\times$ & 12G & 14.09 \\
        \checkmark & \checkmark & 12G & 14.43  \\
        \bottomrule
    \end{tabular}
    \vspace{-5mm}
    \label{tab:ablation2}
\end{table}

\subsection{\textbf{Ablation Study}}
 We provide ablations on SemanticKITTI for the designs of each proposed component. 
 Table \ref{tab:ablation1} demonstrates the effectiveness of each component individually and in combination with other components on the single frame branch. 
In the column of the Distillation Module, (L) or (S) means distilling the temporal branch with the large or small backbone to the single frame branch.

\textbf{2D Semantic Auxiliary Loss}: The effectiveness of 2D Semantic Auxiliary Loss is shown in row 2 of Table \ref{tab:ablation1}. 
It exceeds VoxFormer-S by 0.73 mIoU and scarcely increases GPU memory cost during training. Since auxiliary loss makes it possible to optimize the feature extractor in a shorter path, the performance is largely promoted.

\textbf{Image Conditioned Cross-Attention}: Row 4 of Table \ref{tab:ablation1} shows the advantage of the image-conditioned cross-attention. 
The performance of the framework is improved by 0.35 mIoU with minimal extra memory cost (about 2G). 
Row 5 and row 9 demonstrate that the cross-attention significantly improves the performance of the single frame branch combined with the other two components, by introducing visual cues for completing occluded regions.

\textbf{Scaling-up And Pre-training}: Table \ref{tab:ablation2} shows the positive impacts of scaling-up and pre-training. According to the comparison between row 1 and row 2, it is clear that increasing the parameters of the backbone can significantly improve the performance of occupancy prediction. 
While the results in row 2 and row 3 prove that pre-training the backbone network on driving datasets further improves the performance of occupancy prediction.

\textbf{Distillation Module}: Row 3, 5 and 7 of Table \ref{tab:ablation1} demonstrate the effectiveness of the distillation module. 
Thanks to the transfer of knowledge from the privileged branch, the performance of the single-frame branch significantly increases. 
In addition, by comparing row 6 with row 8 and row 6 with row 9 of Table\ref{tab:ablation1}, it is verified that knowledge from both multiple frames and large models can be introduced to the single-frame branch through the distillation module. 
The comparison between row 8 and row 9 of the table shows that scaling up the backbone of the privileged branch can also enhance the performance of the single-frame branch through the distillation module. 

Table \ref{tab:ablation2} shows the necessity of the distillation module. 
Using a larger backbone results in a significant increase in GPU memory usage during test time (8G $\to$ 12G), while the distillation module can transfer richer knowledge into the single-frame branch at a low cost.
 
\section{Conclusion}
In this paper, we present MonoOcc, a high-performance and efficient framework for monocular semantic occupancy prediction. 
We propose a semantic auxiliary loss and an image-conditioned cross-attention module, improving the existing 3D semantic occupancy prediction method.
By proposing a distillation module to transfer temporal information and richer knowledge to the monocular branch from a privileged branch, we increase the performance of the framework especially on small and long-tailed objects, while striking a balance between performance and efficiency. 
Benefiting from these improvements, MonoOcc achieves SOTA performance on SemanticKITTI benchmark.

\section{Aknowledgement}
This work is supported by the National Natural Science Foundation of China (NSFC) under Grants No.62173324 and the CAS for Grand Challenges under Grants 104GJHZ2022013GC.

\balance
\bibliographystyle{IEEEtran}
\bibliography{ref}
\end{document}